\documentclass{article} % For LaTeX2e
\usepackage{iclr2027_conference,times}

\usepackage{amsmath,amsfonts,bm}

\def\eqref#1{equation~\ref{#1}}
\def\1{\bm{1}}

\DeclareMathAlphabet{\mathsfit}{\encodingdefault}{\sfdefault}{m}{sl}
\SetMathAlphabet{\mathsfit}{bold}{\encodingdefault}{\sfdefault}{bx}{n}

\usepackage{graphicx}
\usepackage{float}
\usepackage{adjustbox}
\usepackage{booktabs}
\usepackage{xcolor}
\usepackage{colortbl}   % \arrayrulecolor -- gray block separators inside the ablations
\usepackage{array}      % centered fixed-width p-column for the model-name columns
\usepackage{tabularx}
\newcolumntype{Y}{>{\centering\arraybackslash}X}
\usepackage{wrapfig}
\usepackage{amsmath,amssymb}
\usepackage{algorithm}
\usepackage{algpseudocode}
\usepackage{hyperref}
\usepackage{url}
\usepackage{xspace}

\makeatletter
\renewcommand{\fps@figure}{!t}
\renewcommand{\fps@table}{!t}
\makeatother

\newcommand{\method}{DailyBrick\xspace}

\newcommand{\assembly}{\mathcal{A}}

\title{DailyBrick: Language-Guided Synthesis of Brick Assembly Instructions}
\author{
Hang Zhang$^{\dagger}$ \quad
Qijian Tian$^{\dagger}$ \quad
Yuan Xie \quad
Xuhong Wang \quad
Xin Tan$^{*}$\\[6pt]
East China Normal University
\quad
Shanghai Jiaotong University\\[6pt]
$^{\dagger}$Equal contribution.
\quad
$^{*}$Corresponding authors.\\
}
\iclrfinalcopy % Uncomment for camera-ready version, but NOT for submission.
\begin{document}
\raggedbottom

\maketitle
\vspace{-6mm}
\begin{figure}[h]
    \centering
    \includegraphics[width=\linewidth]{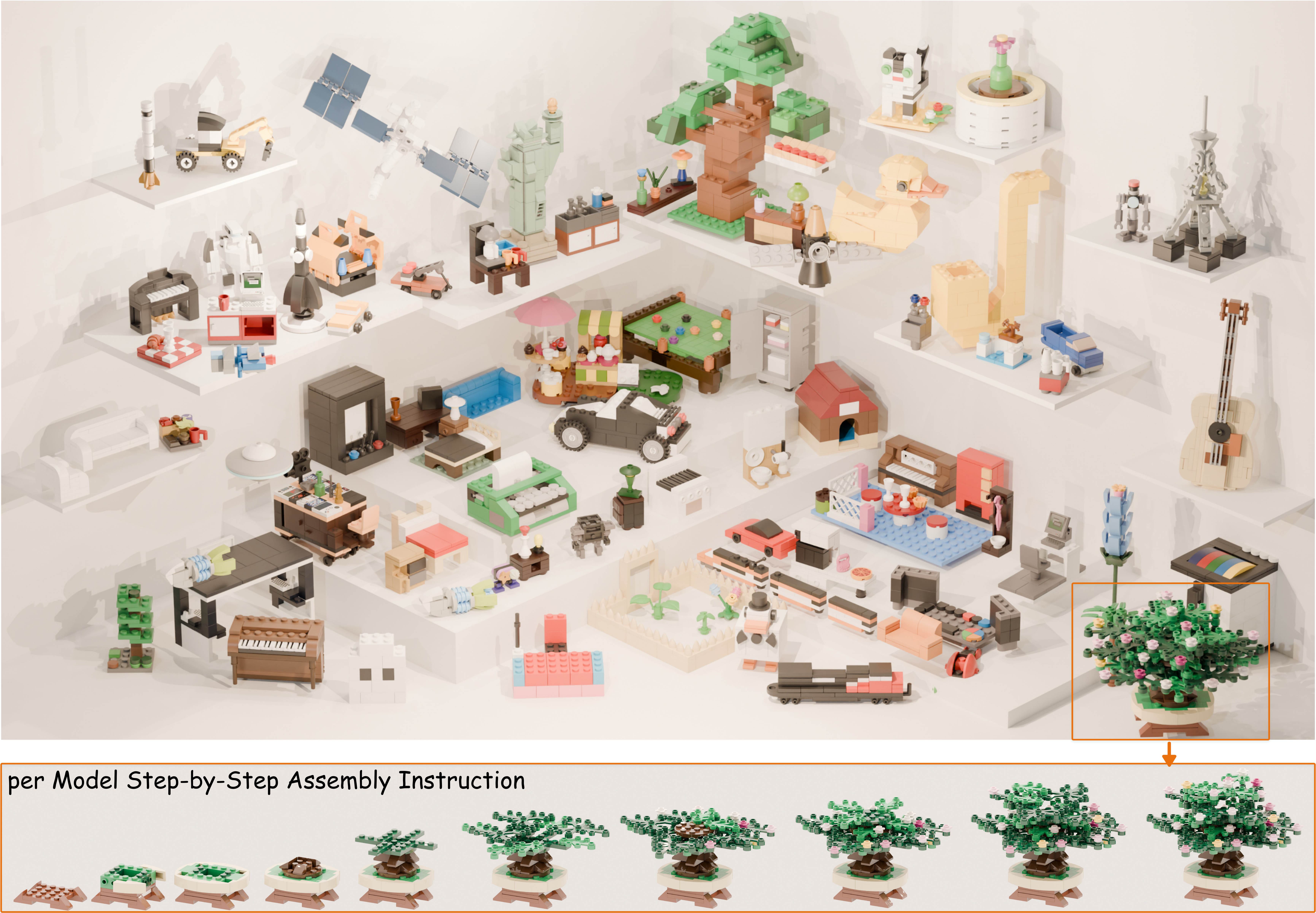}
    \caption{\textbf{LEGO models and assembly instructions generated by DailyBrick.} Given a user request, DailyBrick produces a LEGO model together with step-by-step assembly instructions. We recommend watching the video (in the anonymous link) for a better overview.}
    \label{fig:model-mosaic}
\end{figure}
\vspace{-3mm}

\begin{abstract}
Generating a LEGO model from a user request is more than a plausible 3D shape generation. LEGO models require physical support per brick, and each brick should be placed unblocked. Current methods address this issue by training an expert large language model. We reduce the problem of generating extremely long context on all brick connections to that of writing a program that builds LEGO models like a human, considering each intermediate step could be executed. We introduce \method, a multi-stage agentic system that produces a LEGO model together with a brick-by-brick assembly instruction for human building and robot assembly. We design a multi-stage harness with roles to help LLMs generate more effectively. A versioned construction SDK compiles multi-level decisions, from semantic subparts to individual bricks, into canonical operations. An external Meta-Reviewer agent replays and inspects the generated package, verifies the model and its construction process, identifies failures, traces them to the responsible stage, and routes them to bounded repairs or to editable candidates for later skill or tool updates. Extensive experiments demonstrate that \method generates LEGO models based on simple user requests better than recent agentic systems, and show its utility in downstream applications such as robot manipulations on model assembly from instructions and brick sets (\url{https://spatialagentlab.github.io/DailyBrick.github.io/}).

\end{abstract}

\begin{figure}[t]
    \centering
    \includegraphics[width=0.95\linewidth]{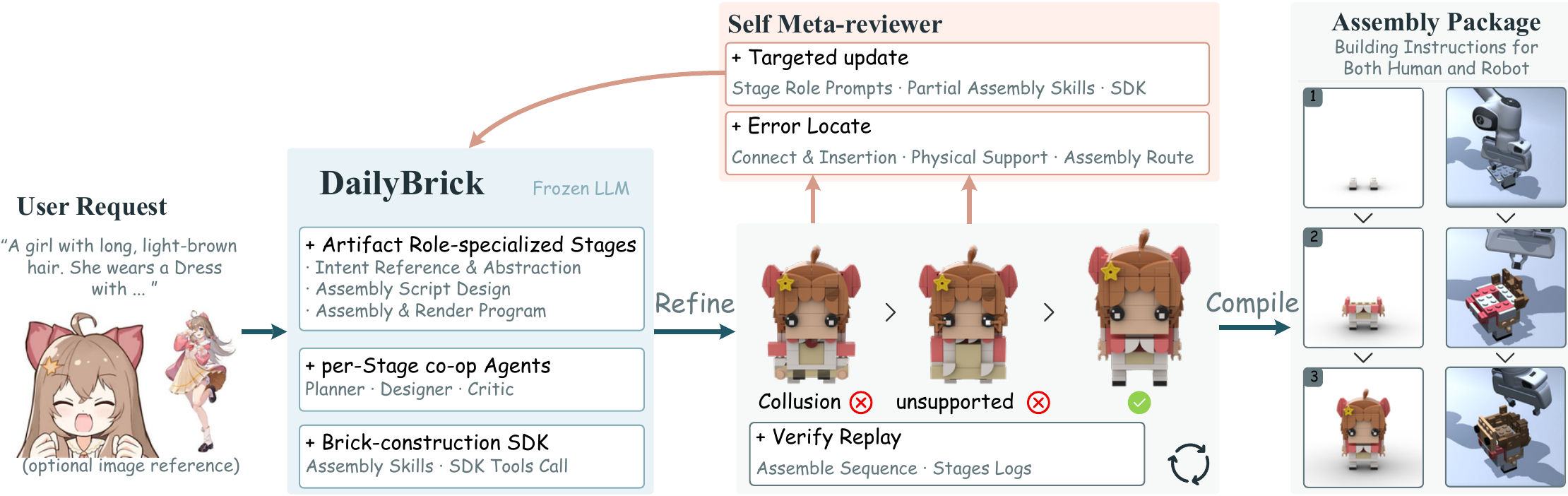}
    \caption{\textbf{System overview.} \method turns a user request into a model, as well as an assembly instruction. Independent prefix replay and an external Meta-Reviewer close the loop by locating failures and routing targeted updates to the responsible stage or SDK component.}
    \label{fig:system-loop}
\end{figure}
\vspace{-8mm}

% intro 单独会画一个对应 contribution 1 的简笔图

\section{Introduction}
% 交代，乐高图纸的效益 对于具身的重要性（参考 具身assembly），去讲这个任务很重要，最大的挑战是缺少从real世界得到对应仿真可逐步复现的大规模多样化资产制作的手段。我们称述这可以 很好的被移植到 乐高图纸生成，因为 图纸 包含 每一行 规定范式的信息、上下文有严格的物理意义、 支持 局部最小单元编改，同时是 自由度非常大 的有边界任务。
Recent advances in 3D generation have made it possible to create diverse object models from text or images. For objects built from multiple parts, however, generating the final geometry does not by itself specify how those parts should be assembled. Assembly connects a desired object to a sequence of actions and intermediate steps, making it a useful setting for studying long-horizon planning, robot manipulation, and error recovery in embodied systems~\citep{sener2022assembly101,heo2023furniturebench}. Generating an object together with its assembly instructions could therefore provide both manuals for a human builder and downstream tasks for robot manipulation. Extending this goal to arbitrary real-world objects is challenging because their parts vary in geometry, connection mechanisms, and physical requirements. LEGO offers a controlled yet expressive setting for studying this problem. Its standardized catalogue defines atomic bricks with fixed geometries and specified connections, while their combinations support diverse objects and structural layouts~\citep{pun2025brickgpt,kulits2026bricknet}. This setting allows us to study the generation of both a model and the brick-by-brick sequence that constructs it.

Even with standardized bricks, a plausible LEGO model does not guarantee an executable assembly instruction. Each step must be compatible with the current partial assembly, making construction depend on both the spatial arrangement of bricks and their placement order. Existing approaches address different aspects of this problem. Image2Lego~\citep{lennon2021image2lego} directly fills a final voxel volume with bricks, while ASAP~\citep{tian2024asap} plans assembly sequences for known parts and target poses through disassembly search. BrickGPT~\citep{pun2025brickgpt} integrates validity checks and physics-aware rollback into autoregressive brick generation. LegoACE~\citep{xu2025legoace} encodes each brick as tokens, yet its authors report that some outputs remain disconnected. BrickNet~\citep{kulits2026bricknet} further represents typed connections and checks construction prefixes for connectivity and collision. These methods advance assembly planning and model generation, but identifying an invalid step does not by itself determine which construction decision should be revised. A failed step may stem from a brick's pose, an incompatible design, or an earlier structural plan. Linking the failure to the decisions that produced it is therefore important for targeted refinement.

%Across existing methods, physical constraints are either applied after a target design is fixed or embedded within generation and sampling. These methods can produce, search, or filter an assembly sequence, but they do not retain an independently replayable record that explains which construction decision caused a failed step and what should be revised. 
Recent agentic workflows offer a way by decomposing long-context generation into structured stages with intermediate outputs and feedback, enabling iterative refinement of generation programs~\citep{yao2022react,hong2023metagpt,wang2024codeact}. General-purpose coding agents such as Claude Code~\citep{claude}, Codex~\citep{codex}, and DeepSeek Harness~\citep{deepseekharness} can plan, write, and iteratively refine software with limited human supervision. ReAct~\citep{yao2022react} and CodeAct~\citep{wang2024codeact} connect language-model reasoning to tool use and executable actions, while MetaGPT~\citep{hong2023metagpt} coordinates specialized agent roles. In 3D generation, SceneSmith~\citep{pfaff2026scenesmith} organizes scene construction into hierarchical stages, and Articraft~\citep{zhou2026articraft} uses executable programs and feedback to refine generated designs. Applying these ideas to brick assembly requires connecting high-level structural decisions to individual placements, so that a failed construction step can be traced back to the decisions that produced it.

%; neither, however, treats the output as a brick-by-brick assembly process whose prefixes must be replayed and routed back to the responsible construction decision.

% 为了验证这样的假设，我们提出了 XXX，一个可以基于 文本描述生成 乐高模型的 agent pipeline。 我们主要基于3点来构建 DailyBrick。DailyBrick 应该是全自动的，给出一个模型的描述，它不需要任何的人为干涉/调整就能更新 内部prompt、skill等，输出 图纸。DailyBrick 应该是可拼接的，等同于任何 3D模型的构建，图纸对应的模型应该是完全物理支撑，不存在浮空、内部碰撞等 manual/机械臂 遵循顺序拼装时无法完成的 序列，它每一步都应该对应图纸的上下两行零件 可以追溯。它应该是有丰富表达的，一项文本描述本就是模糊的，输出在满足语义完整性的情况下每次其实都有差别，同时覆盖大范围 真实物体类别，包括可能的预算不同。 
% 一个关键的技术亮点就是 agent 的设计，我们 DailyBrick 规划了 一系列的stage，从 文本描述 相关的 模型区域规划 到 每一片零件的放置，从而 达到 每一行结果 都可追溯，并且每个 stage 都是独立的，只通过对应格式的输入输出相连（通常为图纸文档）。利用现有 agentic AI， 每一个 stage 可以被认为是 3个独立agent 的多次交互：一个 designer 通过知道 本stage 的任务描述，提出规划；一个 actor 按照规划，读取对应skill或者 工具包（代码patterns），调整 具体细节；以及一个 critic 负责评估 和 loop refinement（出于效率和成本考量，循环次数通常只有几次）。整个过程最后 依赖于一个 meta-reviewer，meta-reviewer在知晓 每个stage 任务安排的情况下，负责评估最后的图纸结果，并基于 输出图纸中 特定文档行的错误，溯源到 发生的stage，以及涉及的 skill 或者工具包。初期严重的时候，meta-reviewer会修改每个stage 具体输入输出格式，甚至stage 本身详细的prompt 安排。但是 每个stage本身的职能 因为天然 自然语言到 模型图纸的过度，而不受到影响。
% eleven typed stages 这个留在方法里说，11个阶段感觉并不是什么值得拿出来说的事情，会让人感觉整个pipeline很复杂，一般复杂的东西并不优雅
% 具体的，SDK是一个特殊且关键的工具，其 的作用在于，链接 LLM 到模型图纸之间，需要专门侧重 和 表达的信息。这样的好处是，让agent可以按照 coding 模型 工作的同时，满足图纸每一步生成的稳定性和可靠性。SDK同时可以 执行 单个零件层级的放置（类型、角度、颜色、位置） 以及 局部部件层级的组合 （规划一个房间），保持 agent pipeline 中每一步的 信息压缩 和 可读性，让 LLM 可以 编写、调用 和验证 模型结构的完整性 和 有效性。
Motivated by these observations, we introduce \method, {\bf an agentic pipeline that turns user requests into LEGO models and their step-by-step assembly instructions}, reviewing each step and locating errors back to target stages or programs that need refinements. \method organizes the pipeline as a sequence of bounded stages, from overall layouts and semantic subpart designs to individual brick placement. A Planner-Designer-Critic workflow with separate contexts at each stage, passing only committed stage outputs to the next. Moreover, a Self Meta-Reviewer uses replay and provenance evidence to route detected defects to the implicated stage and stage output. It decides whether the current package(models and instructions) can be accepted or should trigger an update. Additionally, to bridge agents' semantic planning and executable brick operations on LEGO models, we propose a versioned Software Development Kit (SDK). It supports both individual steps (colours, directions, brick ID) with partial assembly skills, allowing agents to express object-level plans as traceable step-by-step assembly instructions. Moreover, Self Meta-Reviewer replays the instructions to verify the result and locates errors in incomplete replay to update related stage, skills, or tools for refinement.
%贡献点（固定）
% 1. 讲我们的pipeline，text->agent pipeline-> meta-reviewer -> 通过与否 -》最后的图纸结果
% 2. 讲我们设计了一个 可更新的SDK，通过meta0reviewer 定点的在训练集上更新
% 3. 大量实验证实 dailybrick 的性能，可以支撑大规模的数据生成，以及维持同文本的随机生成质量 和 真实世界未见过物体的泛化。
Our contributions are as follows:
\begin{itemize}
  \item We present \method, a multi-stage agentic pipeline that turns user LEGO requests into 3D models and step-by-step assembly instructions, where a Self Meta-Reviewer would locate errors and achieve target updates.
  \item We develop a brick-construction SDK with reusable partial assembly skills and traceable brick operations. Combined with verify replay, these operations provide evidence for the Self Meta-Reviewer to locate errors and guide targeted updates.
  \item Extensive experiments across multiple base models demonstrate that \method could improve LEGO model generation in visual quality and assembly-sequence validity, while simulation examples illustrate downstream assembly and interaction.
\end{itemize}
\vspace{-3mm}
\section{Related Work}
\label{sec:related}
\vspace{-3mm}
\subsection{Assembly generation and process verification}
\vspace{-3mm}
Assembly datasets and benchmarks study recorded human procedures, robotic execution, and sequence planning~\citep{ben2020ikeaasm,sener2022assembly101,heo2023furniturebench,gu2025manualvla,li2026assemblybench}. AssemLM~\citep{jing2026assemlm} combines manuals, point clouds, and text for robotic assembly pose reasoning. ASAP~\citep{tian2024asap} plans assembly from known parts and target poses through disassembly search. For LEGO generation, Image2Lego~\citep{lennon2021image2lego} converts voxel shapes into bricks and instructions. AssemblyComplete~\citep{chen2024assemblycomplete} learns to complete partial LEGO structures while masking physically invalid actions. WorkBenchMark~\citep{ma2026workbenchmark} studies robot assembly with a disassembly baseline. Brick-Composer~\citep{liu2026brickcomposer} studies manual-conditioned brick selection and pose estimation with multimodal models. BrickGPT~\citep{pun2025brickgpt} uses validity checks and physics-aware rollback, whereas BrickAnything~\citep{ni2026brickanything} combines structure-aware tokenization with rollback. LegoACE~\citep{xu2025legoace} generates ordered part tokens, and BrickNet~\citep{kulits2026bricknet} represents typed connections and checks collision. DailyBrick generates the model and assembly instructions from one construction trace, allowing each recorded step to be replayed and located.
\vspace{-3mm}
\subsection{Agentic construction and execution-grounded feedback}
\vspace{-3mm}
ReAct~\citep{yao2022react} and CodeAct~\citep{wang2024codeact} connect language-model reasoning to tool use and executable actions, while MetaGPT~\citep{hong2023metagpt} coordinates agent roles. Self-Refine~\citep{madaan2023selfrefine} and Reflexion~\citep{shinn2023reflexion} use feedback to revise outputs or subsequent behaviour, and Voyager~\citep{wang2023voyager} accumulates reusable skills. In 3D generation, SceneSmith~\citep{pfaff2026scenesmith} uses agent proposal and critique, and Articraft~\citep{zhou2026articraft} combines a coding program with execution feedback. AgentTrace~\citep{wang2026agenttrace} analyzes execution traces to localize failures in multi-agent workflows, while SkillForge~\citep{liu2026skillforge} uses diagnosed failures to refine reusable skills. FAPO~\citep{kassianik2026fapo} optimizes prompts and pipeline structure using execution feedback, while Lin et al.~\citep{lin2026harnessupdating} distinguish the ability to produce harness updates from the ability to benefit from them. DailyBrick uses verified evidence to guide targeted updates: the Self Meta-Reviewer locates errors and identifies the relevant stage outputs, skills, or SDK operations to refine.
\vspace{-1mm}
\begin{figure}[t]
\centering
\includegraphics[width=0.99\linewidth]{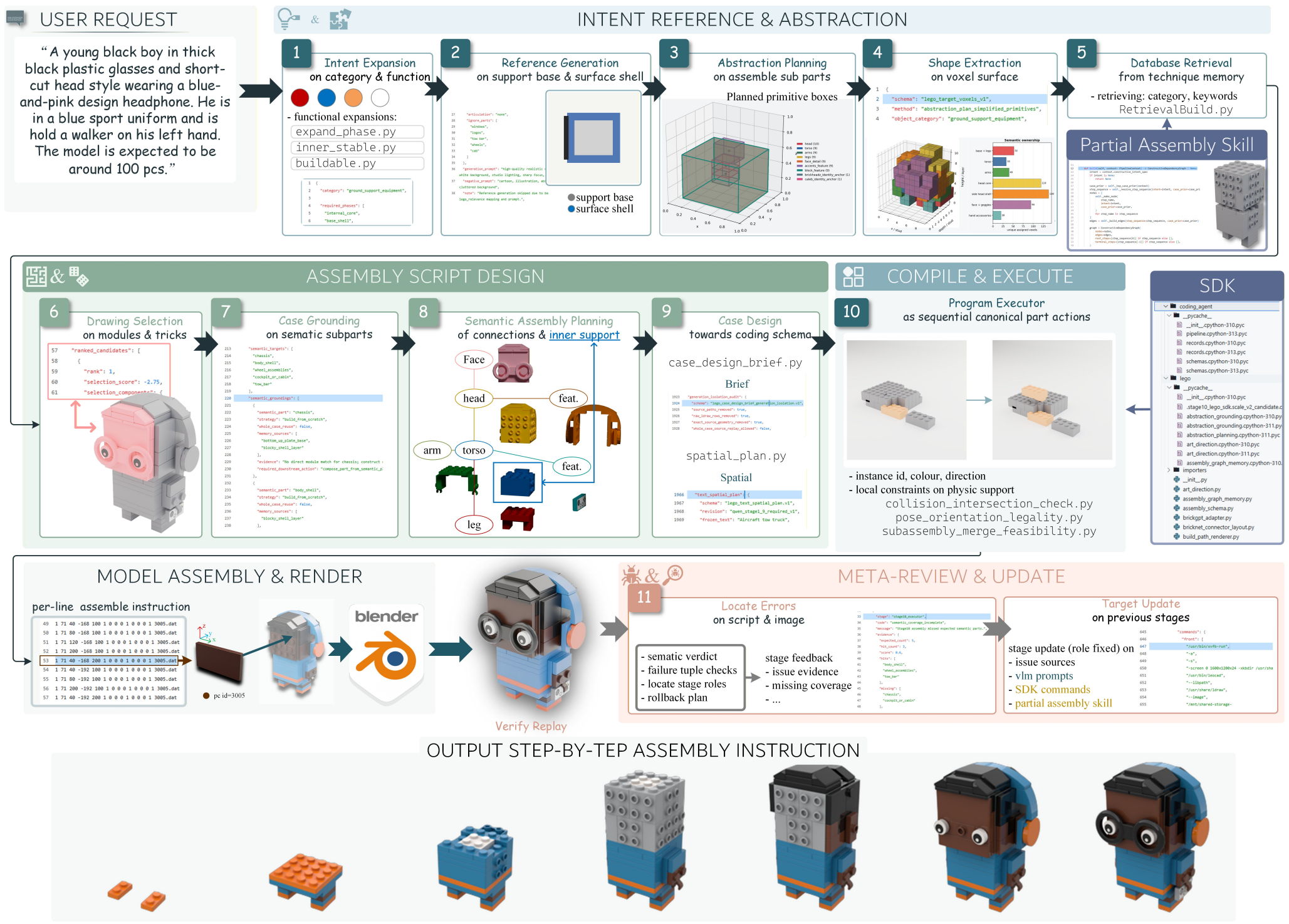}
\caption{\textbf{Overview of \method.} A multi-stage Planner--Designer--Critic workflow uses partial assembly skills and a construction SDK to generate a shared assembly trace, from which the LEGO model and step-by-step instructions are derived. Independent verify replay provides evidence for the Self Meta-Reviewer to locate errors and guide targeted refinements.}
\label{fig:main-pipeline}
\vspace{-3mm}
\end{figure}
\vspace{-3mm}

\section{Method}
\label{sec:method}
\vspace{-3mm}
As shown in Fig.~\ref{fig:main-pipeline}, \method takes a user request \(x\) and produces a LEGO model with step-by-step assembly instructions, or returns a failure report if the required checks cannot be satisfied. The optional image reference would be described as text content for input and verification. Each run uses an updatable configuration \(\Gamma\), comprising P-D-C prompts, partial assembly skills and the SDK.
% 这里可以加个小图，因为很多人其实看了也不清楚
The assembly instruction is represented by an assembly sequence \(\pi=(a_1,\ldots,a_T)\), where each action \(a_t\) specifies a brick placement. The resulting construction package is
\vspace{-2mm}
\begin{equation}
 \mathcal O=(\assembly_T,\pi,G,R),
 \label{eq:delivered-package}
\end{equation}
\vspace{-1mm}
where \(\assembly_T\) is the final LEGO model, \(G=(g_1,\ldots,g_T)\) contains the instruction views, and \(R\) stores verification results and recorded generation dependencies. At step \(t\), the instruction view \(g_t\) shows the placement \(a_t\) that transforms \(\assembly_{t-1}\) into \(\assembly_t\). Each view \(g_t\) illustrates the placement \(a_t\) that transforms the partial assembly \(\assembly_{t-1}\) into \(\assembly_t\). The model and instructions therefore describe the same construction sequence, which also provides part identities and ordered poses for downstream simulation and robot planning. Output package acceptance requires both successful verify replay under physical assembly constraints and a review of whether the generated design satisfies the user request. When a verify replay fails, the resulting evidence is passed to the Self Meta-Reviewer for error localizations and targeted updates.

\vspace{-2mm}
\subsection{Multi-stage Planning and Construction}
\vspace{-2mm}

DailyBrick progressively translates the user request into a construction program through explicit stage outputs:
\vspace{-2mm}
\begin{equation}
 x\longrightarrow \mathcal B\longrightarrow z\longrightarrow q \xrightarrow{\mathrm{Comp}_{\Gamma}}\pi,
 \label{eq:artifact-chain}
\end{equation}
\vspace{-1mm}
where \textsc{DesignBrief} \(\mathcal B\) specifies the request, \textsc{StructuralBlueprint} \(z\) defines support layout and brick connections. \textsc{AssemblyProgram} \(q\) expresses the design through SDK operations. As shown in Fig.~\ref{fig:main-pipeline}, these representations summarize the eleven-stage workflow: early stages interpret the request, middle stages develop the structure and its subparts, and final stages compile and verify the assembly. Each stage receives defined inputs and produces an output checked against its requirements.

Each agent-governed stage follows a Planner--Designer--Critic(PDC) loop. The Planner uses a pre-assigned stage role (fixed) and accepted upstream outputs to propose a local plan based on prompts. The Designer implements the plan within the stage's permitted scope using the available skills and SDK operations. The Critic examines the candidate and requests revisions when needed, while a deterministic reviewer blocks outputs that violate construction requirements and stage roles. The stages use separate contexts and communicate through explicit outputs and feedback on PDC, where dialogues are not passed between stages. Accepted outputs retain their stage identifiers and dependencies, allowing later brick actions to be traced to the decisions that produced them. A refinement creates a new output version while preserving the earlier record.

\vspace{-1mm}
\subsection{Partial Assembly Skills and SDK}
\vspace{-1mm}

The brick-construction SDK connects semantic plans to executable brick operations. Brick-level function calls specify the brick ID, colour, pose, connector, support dependency, and insertion direction. Partial assembly skills compose these calls into reusable procedures for subparts such as a wall, chassis, or roof. Each skill has a defined interface, preconditions, permitted read-write scope, and emitted actions, allowing agents to construct and revise local components within the overall design. 

Agents combine these operations and skills in an \textsc{AssemblyProgram} \(q=(u_1,\ldots,u_L)\). The deterministic compiler \(\mathrm{Comp}_{\Gamma}\) expands the program into the assembly trace \(\pi\), recording the originating program operation for each step. Together, these records link a brick action to the stage output and, where applicable, the partial assembly skill and SDK operation that produced it.

\subsection{Self Meta-Reviewer: Verify Replay and Refinements}
Verify replay checks the assembly trace ($\pi$) from the initial state ($\assembly_0$). Before each placement \(a_t\), it reconstructs the accepted assembly \(\assembly_{t-1}\) and its attachment frontier \(\mathcal F_{t-1}\), which records the attachment sites available for subsequent placements. It then checks the proposed step against physical constraints in ($z$):
\vspace{-1mm}
\begin{equation}
 \kappa_t=\mathrm{CheckTransition}_{\Gamma}
 (\assembly_{t-1},\mathcal F_{t-1},a_t,z),
 \label{eq:prefix-check}
\end{equation}
\vspace{-1mm}
where \(\kappa_t\) records the result of brick pose validity, connector compatibility and subpart assembly connections. A passing placement produces \(\assembly_t\) and updates the frontier. A rejected step stops construction replay and produces a failure report containing the action, violated constraint, and last accepted state.

The Self Meta-Reviewer uses this report and recorded generation dependencies to locate errors and select candidate revision targets. Starting from the rejected step, it traces the associated program operation, stage output, and relevant skill or SDK operation.  The violated constraint and dependency links guide the choice of what to revise and which dependent outputs must be re-verified or refined. If the available evidence does not support a revision target, the Self Meta-Reviewer returns \texttt{defer}.

Refinement operates at two levels. Within a run, targeted updates modify the affected construction program or stage outputs while keeping the configuration \(\Gamma\) fixed. The revised program is recompiled and undergoes complete verify replay before acceptance. Across completed runs, recurring errors can motivate candidate updates to stage prompts, partial assembly skills, or SDK operations. These candidates are evaluated on the triggering cases and protected cases before being activated in a later configuration.
\begin{figure}[t]
\centering
\includegraphics[width=\linewidth]{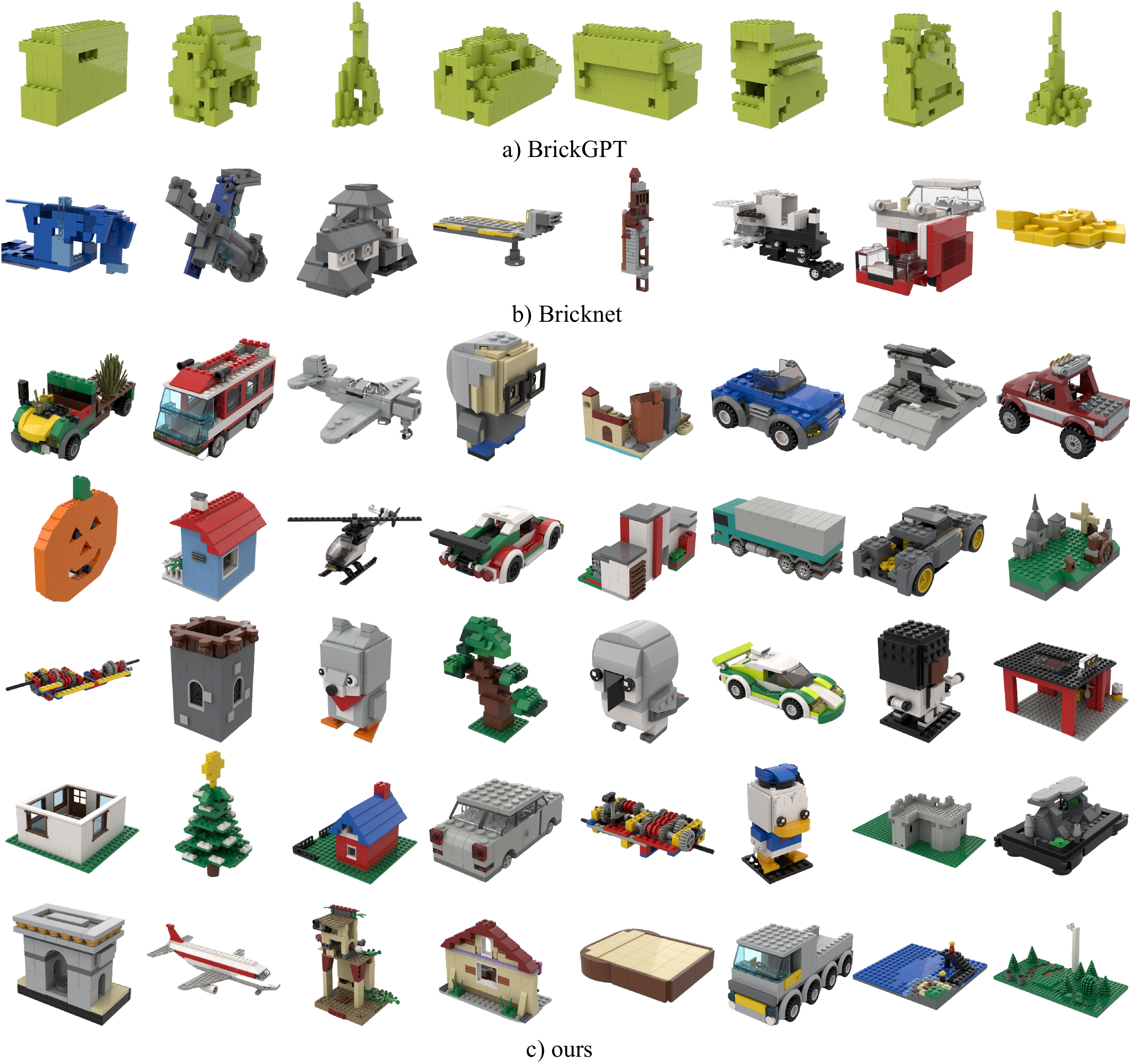}
\caption{\textbf{Generated Examples.} Rows (a)--(c) show randomly outputs from BrickGPT~\citep{pun2025brickgpt}, BrickNet~\citep{kulits2026bricknet}, and \method, respectively.}
\label{fig:qualitative-comparison}
\end{figure}
\vspace{-3mm}

\section{Experiments}
\label{sec:experiments}
\subsection{Experimental Setup}
% Requires \usepackage{booktabs,graphicx}

\begin{wraptable}{r}{0.5\textwidth}
\centering
\vspace{-6mm}
\caption{\textbf{Dataset Statistics.} In contrast to BrickGPT~\citep{pun2025brickgpt} and BrickNet~\citep{kulits2026bricknet}, we increase the scale and part diversity of models.}
\label{tab:brick_dataset_scale}
\scriptsize
\setlength{\tabcolsep}{2.0pt}
\renewcommand{\arraystretch}{1.08}
\begin{adjustbox}{max width=\linewidth}
\begin{tabular}{@{}lrrrr@{}}
\toprule
Dataset & \#Models & Avg. & Max & \#Types \\
\midrule
BrickGPT~\citep{pun2025brickgpt} & 47{,}389 & 107.56 & 409 & 8 \\
BrickNet~\citep{kulits2026bricknet}  & 321{,}320 & 126.24 & 131{,}632 & 9{,}746 \\
\textbf{Ours} & \textbf{358{,}775} & \textbf{172.58} & \textbf{259{,}179} & \textbf{25{,}406} \\
\bottomrule
\end{tabular}
\end{adjustbox}
\vspace{-3mm}
\end{wraptable}

\paragraph{Dataset.}
We evaluate on a self-collected dataset of 358,775 structured LEGO models, including a fixed test split of 18,653 cases. Each model is associated with a text description and brick-count information and is represented using the public LDraw format and part library~\citep{ldraw_file_format}. Table~\ref{tab:brick_dataset_scale} compares the dataset with the corpora reported by BrickGPT and BrickNet~\citep{pun2025brickgpt,kulits2026bricknet}. These statistics describe dataset scale and part diversity rather than a matched training-data comparison.

\vspace{-1mm}

\paragraph{Baselines.}
We compare with direct pose generation and public implementations of BrickGPT and BrickNet~\citep{pun2025brickgpt,kulits2026bricknet}. We also evaluate GPT 6-Astra~\citep{openai2026gpt6astra} as a text-to-brick baseline and pair it with the DailyBrick framework. 

\paragraph{Implementation Details.}
The main pipeline uses Qwen3.6-27B~\citep{qwen2026qwen36_27b}. Planner, Designer, and Critic are assigned as agents with individual context. Each stage permits three role rounds, and a request permits three attempts. A frozen case manifest records the request, inventory, seed, model and prompt versions, role configuration, tool and verifier versions, budgets, serialized artifacts, transition reports, and package decision. Invalid generations remain in the denominator and receive zero for success metrics.

\begin{figure}[ht]
\centering
\includegraphics[width=\linewidth]{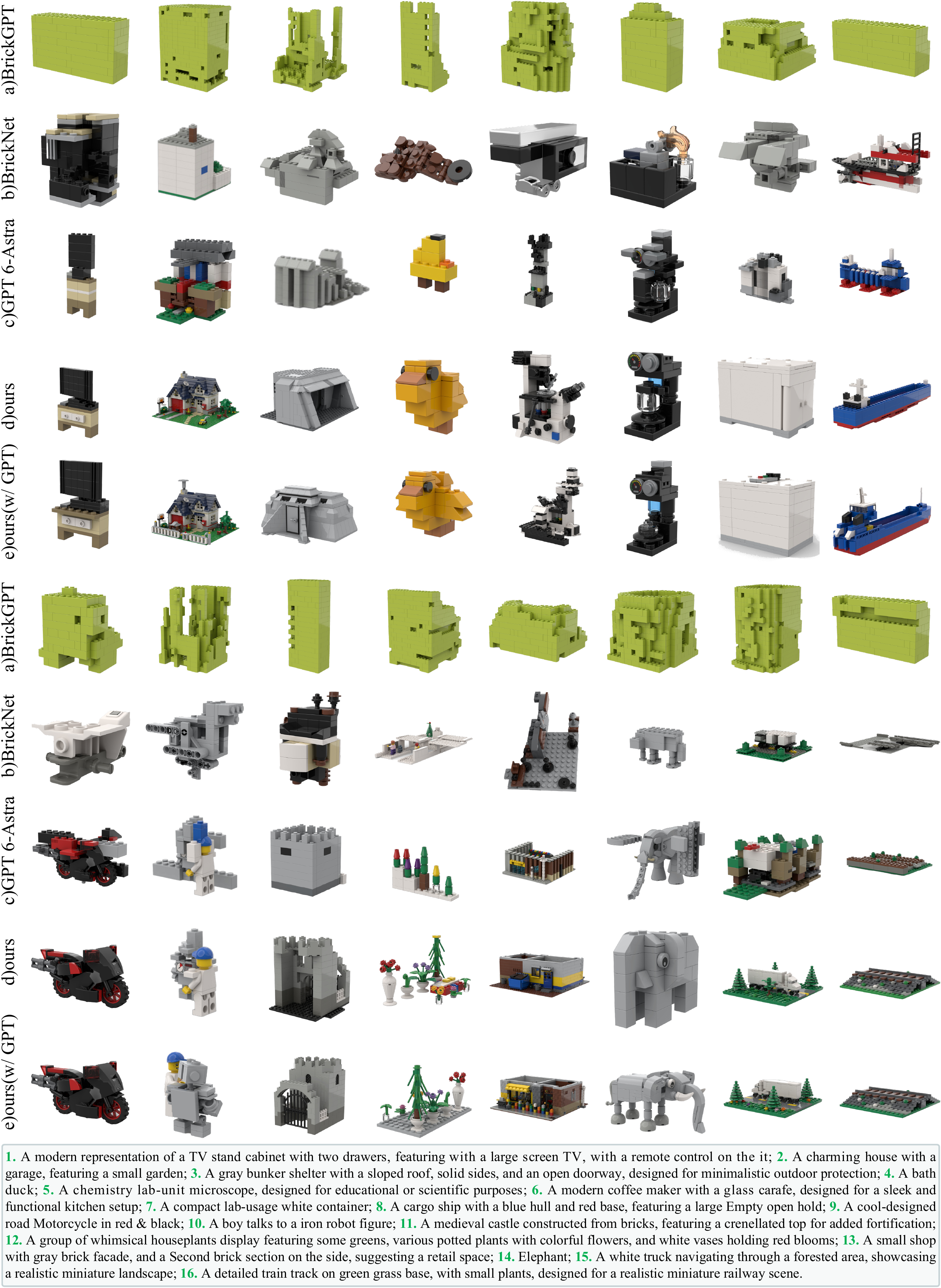}
\vspace{-6mm}
\caption{\textbf{Text-conditioned generation results.} Outputs for 16 evaluation prompts are arranged in two blocks of eight columns. Within each block, BrickGPT~\citep{pun2025brickgpt}, BrickNet~\citep{kulits2026bricknet}, GPT 6-Astra~\citep{openai2026gpt6astra} and \method appear from top to bottom. Each column corresponds to the numbered prompt below.}
\label{fig:matched-comparison}
\vspace{-3mm}
\end{figure}

\begin{table}[t]
\centering
\caption{\textbf{Main Results on Text-Conditioned Generation}. We compare BrickGPT~\citep{pun2025brickgpt}, BrickNet~\citep{kulits2026bricknet}, GPT 6-Astra, and DailyBrick frameworks in visual quality, text alignment, and structural validity. Baseline scores combine published results and additional measurements under their respective evaluation protocols. The best value in each column is shown in \textbf{bold}.}
\label{tab:main-comparison}
\scriptsize
\setlength{\tabcolsep}{2.5pt}
\renewcommand{\arraystretch}{1.12}
\begin{adjustbox}{max width=\linewidth}
\begin{tabular}{@{}l*{9}{c}@{}}
\toprule
Method & Valid (\%)\(\uparrow\) & Stable (\%)\(\uparrow\) & \shortstack{CLIP $\uparrow$\\{\tiny\citep{radford2021clip}}} & CLIP-L/14 \(\uparrow\) & \shortstack{DINOv2 $\uparrow$\\{\tiny\citep{oquab2024DINOv2}}}
& \(P_{\mathrm{inv}}\) \(\downarrow\) & \shortstack{VQAScore $\uparrow$\\{\tiny\citep{lin2024vqascore}}} & \shortstack{PE $\uparrow$\\{\tiny\citep{bolya2025perceptionencoder}}} & \shortstack{SigLIP2 $\uparrow$\\{\tiny\citep{tschannen2025siglip2}}} \\
\midrule
BrickGPT~\citep{pun2025brickgpt} & 100.0 & 97.91  & 0.322 & 20.14 & 0.877 & 0.064 & 0.048 & 0.157 & 0.050 \\
BrickNet (14B)~\citep{kulits2026bricknet}   & 96.68    & 94.30  & 0.339 & 22.93 & 0.920 & 0.228 & 0.600 & 0.282 & 0.623 \\
GPT 6-Astra~\citep{openai2026gpt6astra}  & 100.0 & 100.0 & 0.330 & 26.54 & 0.895 & 0.057 & 0.625 & 0.297 & 0.626 \\
DailyBrick (w/ Qwen3.6-27B) & 100.0 & 100.0 & 0.340 & 27.85 & 0.906 & 0.049 & 0.736 & 0.362 & 0.724 \\
DailyBrick (w/ GPT 6-Astra) & 100.0 & 100.0 & \textbf{0.358} & \textbf{31.26} & \textbf{0.923} & \textbf{0.044} & \textbf{0.767} & \textbf{0.363} & \textbf{0.734} \\
\bottomrule
\end{tabular}
\end{adjustbox}
\end{table}

\begin{table}[!t]
\centering
\caption{\textbf{Efficiency and Cost Analysis.}
Results are reported on the 18,653-case test split. Tokens and time are measured per brick among different model scales. A dash indicates that repair success is not applicable. The best value in each column is \textbf{bold}.}
\label{tab:efficiency-analysis}

\scriptsize
\setlength{\tabcolsep}{2.5pt}
\renewcommand{\arraystretch}{1.08}

\begin{tabular*}{\linewidth}{@{\extracolsep{\fill}}l*{5}{c}@{}}
\toprule
Method
& \shortstack{Tokens/case $\downarrow$}
& \shortstack{Time (s) $\downarrow$}
& \shortstack{Prefix pass $\uparrow$}
& \shortstack{Repair success $\uparrow$}
& \shortstack{Accept $\uparrow$} \\
\midrule

BrickGPT~\citep{pun2025brickgpt}
& \textbf{2.312k} & 65.62 & 0.706 & -- & 0.684 \\

BrickNet (14B)~\citep{kulits2026bricknet}
& 3.544k & 56.68 & 0.690 & -- & 0.743 \\

GPT 6-Astra~\citep{openai2026gpt6astra}
& 16.24k & 324.6 & 0.884 & 0.525 & 0.714 \\

DailyBrick (w/ Qwen3.6-27B)
& 7.760k & 207.7 & 0.960 & 0.576 & 0.834 \\

DailyBrick (w/ GPT 6-Astra)
& 4.346k & \textbf{49.09} & \textbf{0.974}
& \textbf{0.677} & \textbf{0.885} \\

\bottomrule
\end{tabular*}
\end{table}

\subsection{Main Results}
\label{sec:main-results}
Table~\ref{tab:main-comparison} evaluates final-model structural validity, visual similarity, and request alignment. Figure~\ref{fig:matched-comparison} presents outputs for the same text prompts, allowing qualitative comparison of object interpretation, component layout, and structural detail. Table~\ref{tab:process-comparison} evaluates the corresponding construction sequences through independent verify replay, measuring agreement with recorded outcomes and states as well as whether all placements pass the construction checks.

\paragraph{Efficiency Analysis.}
\label{sec:efficiency-analysis}
Table~\ref{tab:efficiency-analysis} reports token use and wall-clock time per case alongside construction outcomes, relating generation cost to sequence validity and package acceptance. Prefix pass measures the fraction of cases whose first generated sequence passes all construction checks, whereas Accept measures final package acceptance after the allowed retries. Repair success measures the fraction of repair attempts that pass the required checks.

% Requires: booktabs, graphicx, adjustbox
\begin{table*}[t]
\centering
\caption{\textbf{Component ablation of DailyBrick}. Base denotes zero-shot Qwen3.6-27B. Components are added cumulatively: (i) Planner-Designer-Critic collaboration; (ii) the construction SDK and partial assembly skills; and (iii) the Self Meta-Reviewer. Results are reported on the 18,653-case test split. Replay exact, all-prefix pass, and package pass rate are reported in percentages. Failure position is normalized by sequence length. The best values in each column are bold.}
\label{tab:contribution-ablation}

\scriptsize
\setlength{\tabcolsep}{2.4pt}
\renewcommand{\arraystretch}{1.15}

\begin{adjustbox}{max width=\linewidth}
\begin{tabular}{@{}l*{13}{c}@{}}
\toprule
Variant
& \shortstack{Valid (\%)$\uparrow$}
& \shortstack{Stable (\%)$\uparrow$}
& \shortstack{CLIP$\uparrow$}
& \shortstack{DINOv2$\uparrow$}
& \shortstack{$P_{\mathrm{inv}}$$\downarrow$}
& \shortstack{VQAScore$\uparrow$}
& \shortstack{PE$\uparrow$}
& \shortstack{SigLIP2$\uparrow$}
& \shortstack{Replay exact (\%)$\uparrow$}
& \shortstack{All-prefix pass$\uparrow$}
& \shortstack{Failure pos.$\uparrow$}
& \shortstack{Accept$\uparrow$}
& \shortstack{Tokens/case$\downarrow$} \\
\midrule

Base
& 87.92 & 82.93 & 0.310 & 0.864
& 0.214 & 0.567 & 0.347 & 0.602
& 90.57 & 0.888 & 0.904 & 0.563 & \textbf{5.233k} \\

Base + (i)
& 88.17 & 88.04 & 0.339 & 0.897
& 0.201 & 0.721 & 0.360 & 0.720
& 94.58 & 0.936 & 0.935 & 0.626 & 6.175k \\

Base + (i) + (ii)
& \textbf{100.0} & 99.84 & 0.339 & 0.897
& 0.049 & 0.721 & 0.360 & 0.720
& 95.62 & 0.943 & 0.949 & 0.676 & 7.253k \\

DailyBrick(\textbf{full})
& \textbf{100.0} & \textbf{100.0}
& \textbf{0.340} & \textbf{0.906}
& \textbf{0.049} & \textbf{0.736}
& \textbf{0.362} & \textbf{0.724}
& \textbf{100.0} & \textbf{1.0}
& \textbf{0.960} & \textbf{0.834} & 7.760k \\
\bottomrule
\end{tabular}
\end{adjustbox}
\end{table*}
% 导言区需要：
% \usepackage{wrapfig,booktabs}

\begin{wraptable}{r}{0.50\textwidth}
\centering
\vspace{-12mm}
\caption{\textbf{Assembly Instruction Evaluation.} Results are reported on the full test split. Replay exact (\%) is the fraction of traces whose independent replay matches every recorded step outcome and state, including failures. All-prefix pass (\%) is the fraction of sequences with no placement rejected by verify replay. Failure position is the percentage of the first rejected placement, with successful sequences assigned as 1. Best values are bold.}
\label{tab:process-comparison}

\scriptsize
\setlength{\tabcolsep}{2pt}
\renewcommand{\arraystretch}{1.08}

\begin{adjustbox}{max width=\linewidth}
\begin{tabular}{@{}l*{3}{r}@{}}
\toprule
Method
& \shortstack{Replay exact (\%)$\uparrow$}
& \shortstack{All-prefix pass $\uparrow$}
& \shortstack{Failure pos. $\uparrow$} \\
\midrule
BrickGPT~\citep{pun2025brickgpt}
& \textbf{100.0} & 0.706 & 0.98 \\
BrickNet (14B)~\citep{kulits2026bricknet}
& 96.61 & 0.690 & 0.43 \\
GPT 6-Astra~\citep{openai2026gpt6astra}
& 96.02 & 0.884 & 0.44 \\
DailyBrick (Qwen3.6-27B)
& \textbf{100.0} & 0.960 & \textbf{1} \\
DailyBrick (GPT 6-Astra)
& \textbf{100.0} & \textbf{0.974} & \textbf{1} \\
\bottomrule
\end{tabular}
\end{adjustbox}
\end{wraptable}
\vspace{-3mm}
% Requires: booktabs, adjustbox
\begin{table}[t]
\centering
\caption{\textbf{Base-model ablation.}
Each pair compares single-shot generation by a base model
with DailyBrick using the same model, denoted by Ours.
Results are reported on the test split.}
\label{tab:base-model-ablation}

\scriptsize
\setlength{\tabcolsep}{2.4pt}
\renewcommand{\arraystretch}{1.12}

\begin{adjustbox}{max width=\linewidth}
\begin{tabular}{@{}l*{10}{c}@{}}
\toprule
Method
& \shortstack{CLIP$\uparrow$}
& \shortstack{DINOv2$\uparrow$}
& \shortstack{$P_{\mathrm{inv}}$$\downarrow$}
& \shortstack{VQAScore$\uparrow$}
& \shortstack{SigLIP2$\uparrow$}
& \shortstack{Replay exact (\%)$\uparrow$}
& \shortstack{All-prefix pass$\uparrow$}
& \shortstack{Failure pos.$\uparrow$}
& \shortstack{Accept$\uparrow$}
& \shortstack{Tokens/case$\downarrow$} \\
\midrule

GPT-4o
& 0.306 & 0.861 & 0.298 & 0.543 & 0.584
& 89.7 & 0.871 & 0.892 & 0.542 & 6.78k \\

Ours (w/ GPT-4o)
& 0.339 & 0.903 & 0.0492 & 0.728 & 0.722
& 96.0 & 0.958 & 0.960 & 0.824 & 7.90k \\

\midrule

Qwen3.6-27B
& 0.310 & 0.864 & 0.214 & 0.567 & 0.602
& 90.6 & 0.888 & 0.904 & 0.563 & 5.23k \\

Ours (w/ Qwen3.6-27B)
& 0.340 & 0.906 & 0.0490 & 0.736 & 0.724
& \textbf{100} & 0.960 & \textbf{1.00} & 0.834 & 7.76k \\

\midrule

Qwen3.6-35B-A3B
& 0.316 & 0.873 & 0.248 & 0.599 & 0.611
& 91.6 & 0.900 & 0.912 & 0.596 & 4.97k \\

Ours (w/ Qwen3.6-35B-A3B)
& 0.343 & 0.911 & 0.0479 & 0.756 & 0.727
& \textbf{100} & 0.960 & \textbf{1.00} & 0.846 & 7.17k \\

\midrule

Qwen3.8-27B
& 0.324 & 0.885 & 0.166 & 0.601 & 0.622
& 92.7 & 0.915 & 0.927 & 0.624 & 4.75k \\

Ours (w/ Qwen3.8-27B)
& 0.348 & 0.917 & 0.0465 & 0.758 & 0.728
& \textbf{100} & 0.965 & \textbf{1.00} & 0.852 & 7.17k \\

\midrule

Gemini-3 Pro
& 0.327 & 0.886 & 0.106 & 0.619 & 0.624
& 95.3 & 0.946 & 0.947 & 0.664 & 4.99k \\

Ours (w/ Gemini-3 Pro)
& 0.357 & 0.917 & 0.0452 & 0.760 & \textbf{0.734}
& \textbf{100} & 0.971 & \textbf{1.00} & 0.861 & 6.57k \\

\midrule

GPT 6-Astra
& 0.330 & 0.895 & 0.0570 & 0.625 & 0.626
& 96.0 & 0.884 & 0.440 & 0.714 & 16.2k \\

Ours (w/ GPT 6-Astra)
& \textbf{0.358} & \textbf{0.923} & \textbf{0.0440}
& \textbf{0.767} & \textbf{0.734}
& \textbf{100} & \textbf{0.974} & \textbf{1.00}
& \textbf{0.885} & \textbf{4.35k} \\

\bottomrule
\end{tabular}
\end{adjustbox}
\end{table}

\subsection{Ablation Study}
\label{sec:ablation}
Table~\ref{tab:contribution-ablation} evaluates the cumulative addition of Planner--Designer--Critic collaboration, the construction SDK with partial assembly skills, and the Self Meta-Reviewer to single-shot Qwen3.6-27B. This comparison assesses how the components affect model quality, construction validity, and generation cost. Table~\ref{tab:base-model-ablation} pairs single-shot generation with DailyBrick for each base model using the same requests and part inventory, testing whether the framework's benefits extend across base models.

\begin{figure}[t]
\centering
\includegraphics[width=\linewidth]{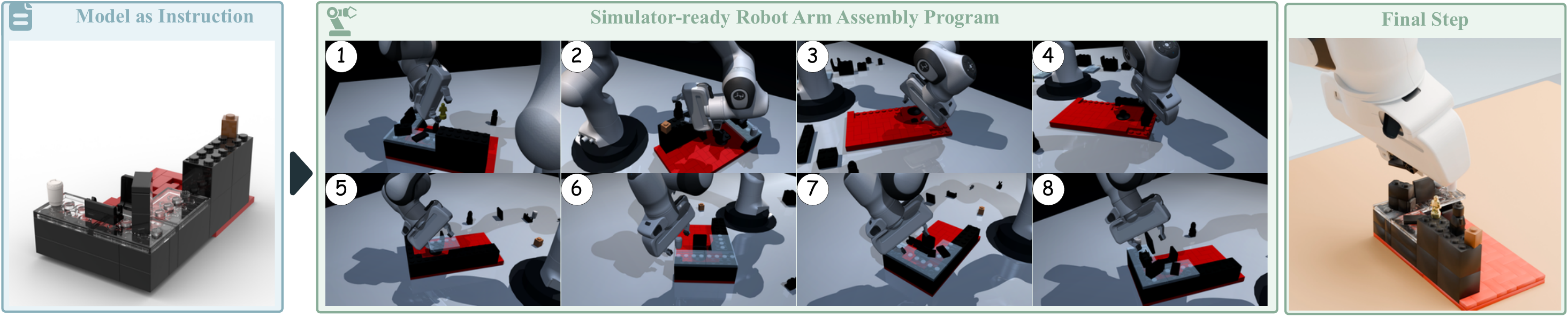}
\caption{\textbf{Step-by-Step Assembly in Simulation.} Left: the generated model. Middle: eight snapshots of a simulated manipulator following the assembly instructions. Right: the manipulator holding the final brick before placement.}
\label{fig:robot-simulation}
\end{figure}

\begin{figure*}[t]
\centering
\begin{minipage}[h]{0.49\textwidth}
\centering
\includegraphics[width=\linewidth]{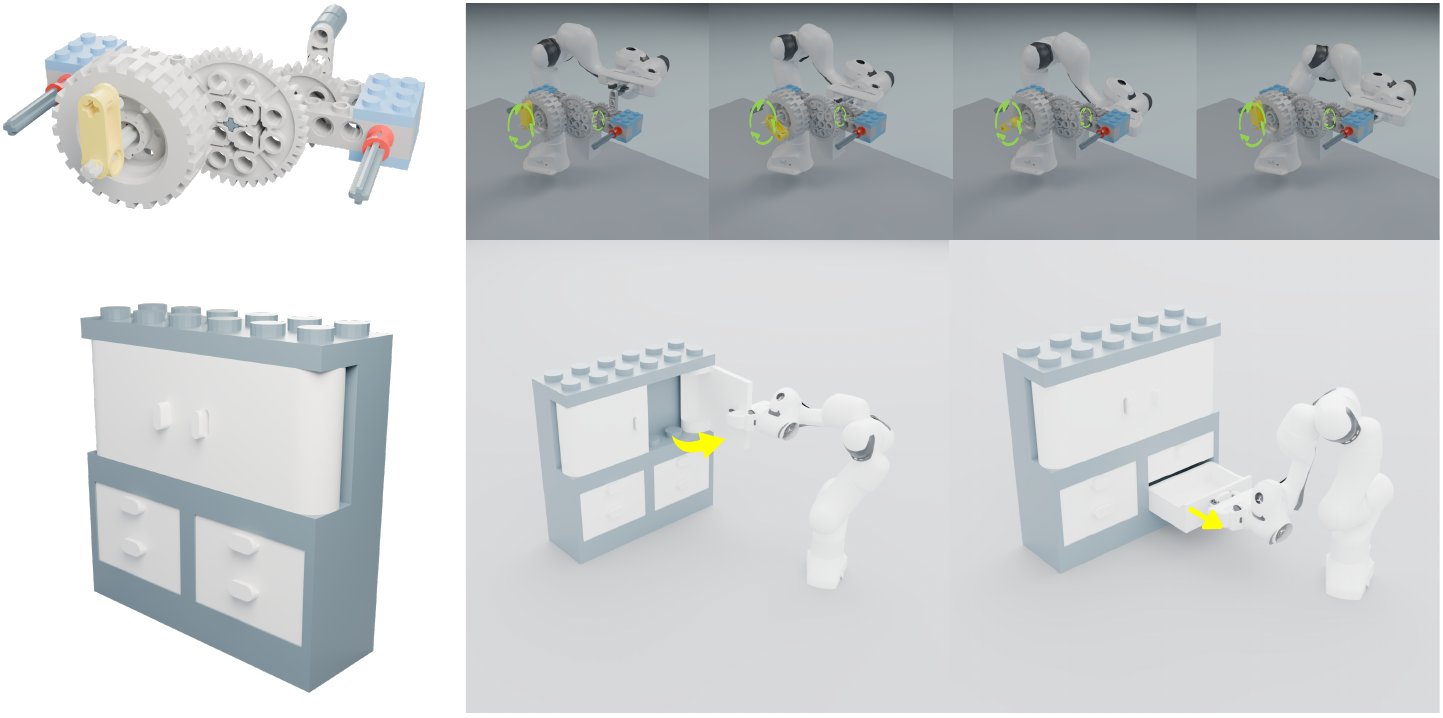}
\caption{\textbf{Physical Interaction in Simulation.} The snapshots show a robot arm could interact with DailyBrick-generated models (rotating \& articulated).}
\label{fig:robot-arm-simulation}
\end{minipage}
\begin{minipage}[h]{0.49\textwidth}
\centering
\includegraphics[width=\linewidth]{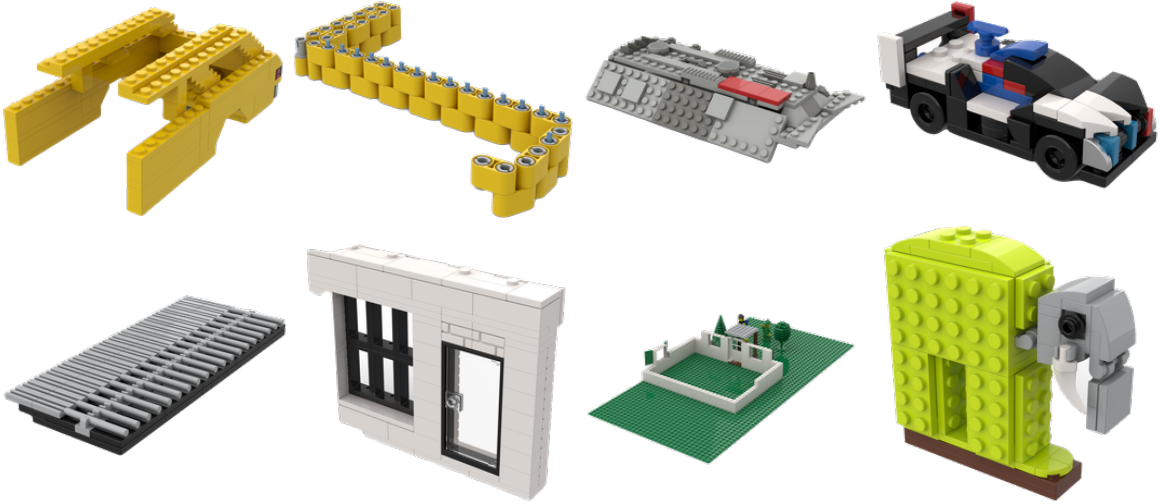}
\caption{\textbf{Limitation.}  Models with valid assembly instructions can still exhibit inconsistent color choices, excessively repetitive structures, and unintended asymmetry.}
\label{fig:failure-cases}
\end{minipage}
\end{figure*}

\subsection{Assembly Execution and Failure Cases}
\label{sec:repair-visual-analysis}
Figures~\ref{fig:robot-simulation} and~\ref{fig:robot-arm-simulation} illustrate two uses of DailyBrick outputs in simulation. The first shows a simulated manipulator following the generated part identities and ordered poses to assemble a model. The second shows interaction with rotating and articulated components of generated models. Together, these examples illustrate how the construction outputs can support both assembly and subsequent interaction in a simulator.

\section{Conclusion}
\label{sec:conclusion}

We presented DailyBrick, an agentic framework that jointly generates brick models and step-by-step assembly instructions
from user requests. By treating the construction sequence as an explicit output, DailyBrick maintains consistency among the final model, intermediate assembly states, and instructions through a shared assembly trace. The framework combines staged planning, partial assembly skills, and a construction SDK with independent verify replays, while recorded dependencies allow the Self Meta-Reviewer to locate errors and guide targeted updates. Experiments across multiple base models show improvements over single-shot generation in visual quality, text alignment, and assembly-sequence validity, supporting the value of jointly generating and checking models and their construction processes. Despite these gains, construction validity does not guarantee design quality. As shown in Figure~\ref{fig:failure-cases}, valid assemblies can still exhibit inconsistent colour choices, excessively repetitive structures, and unintended asymmetry.
\bibliography{references}

@misc{ldraw_file_format,
  author       = {{LDraw.org Standards Board}},
  title        = {{LDraw File Format Specification}},
  year         = {2012},
  url          = {https://www.ldraw.org/article/218.html},
  note         = {Revision 1.0.2, 29 April 2012. Accessed: 2026-09-26}
}

@InProceedings{pun2025brickgpt,
  author       = {Pun, Ava and Deng, Kangle and Liu, Ruixuan and Ramanan, Deva and Liu, Changliu and Zhu, Jun-Yan},
  title        = {Generating Physically Stable and Buildable Brick Structures from Text},
  year         = {2025},
  booktitle    = {Proceedings of the IEEE/CVF International Conference on Computer Vision},
  pages        = {14798--14809},
  doi          = {10.1109/ICCV51701.2025.01373},
  eprint       = {2505.05469},
  archivePrefix = {arXiv},
  url          = {https://doi.org/10.1109/ICCV51701.2025.01373}
}

@InProceedings{kulits2026bricknet,
  author       = {Kulits, Peter and Schmid, Cordelia},
  title        = {{BrickNet}: Graph-Backed Generative Brick Assembly},
  year         = {2026},
  booktitle    = {Proceedings of the IEEE/CVF Conference on Computer Vision and Pattern Recognition (CVPR)},
  eprint       = {2604.22984},
  archivePrefix = {arXiv},
  url          = {https://arxiv.org/abs/2604.22984}
}

@misc{lennon2021image2lego,
  author       = {Lennon, Kyle and Fransen, Katharina and O'Brien, Alexander and Cao, Yumeng and Beveridge, Matthew and Arefeen, Yamin and Singh, Nikhil and Drori, Iddo},
  title        = {{Image2Lego}: Customized {LEGO} Set Generation from Images},
  year         = {2021},
  eprint       = {2108.08477},
  archivePrefix = {arXiv},
  url          = {https://arxiv.org/abs/2108.08477}
}

@InProceedings{tian2024asap,
  author       = {Tian, Yunsheng and Willis, Karl D. D. and Al Omari, Bassel and Luo, Jieliang and Ma, Pingchuan and Li, Yichen and Javid, Farhad and Gu, Edward and Jacob, Joshua and Sueda, Shinjiro and Li, Hui and Chitta, Sachin and Matusik, Wojciech},
  title        = {{ASAP}: Automated Sequence Planning for Complex Robotic Assembly with Physical Feasibility},
  year         = {2024},
  booktitle    = {Proceedings of the IEEE International Conference on Robotics and Automation},
  pages        = {4380--4386},
  doi          = {10.1109/ICRA57147.2024.10611595},
  eprint       = {2309.16909},
  archivePrefix = {arXiv},
  url          = {https://doi.org/10.1109/ICRA57147.2024.10611595}
}

@InProceedings{xu2025legoace,
  author       = {Xu, Hao and Zhang, Yuqing and Wu, Yiqian and Zheng, Xinyang and Liu, Yutao and Tang, Xiangjun and Yang, Yunhan and Liang, Ding and Liu, Yingtian and Guo, Yuanchen and Cao, Yanpei and Jin, Xiaogang},
  title        = {{LegoACE}: Autoregressive Construction Engine for Expressive {LEGO} Assemblies},
  year         = {2025},
  booktitle    = {SIGGRAPH Asia 2025 Conference Papers},
  publisher    = {ACM},
  doi          = {10.1145/3757377.3763881},
  url          = {https://doi.org/10.1145/3757377.3763881}
}

@InProceedings{li2026assemblybench,
  author       = {Li, Danrui and Zhang, Jiahao and Egger, Bernhard and Chatterjee, Moitreya and Lohit, Suhas and Marks, Tim K. and Cherian, Anoop},
  title        = {{AssemblyBench}: Physics-Aware Assembly of Complex Industrial Objects},
  year         = {2026},
  booktitle    = {Proceedings of the IEEE/CVF Conference on Computer Vision and Pattern Recognition (CVPR)},
  url          = {https://merl.com/research/highlights/assemblybench}
}

@misc{ben2020ikeaasm,
  author       = {Ben-Shabat, Yizhak and Yu, Xin and Saleh, Fatemeh Sadat and Campbell, Dylan and Rodriguez-Opazo, Cristian and Li, Hongdong and Gould, Stephen},
  title        = {The {IKEA ASM} Dataset: Understanding People Assembling Furniture through Actions, Objects and Pose},
  year         = {2020},
  eprint       = {2007.00394},
  archivePrefix = {arXiv},
  url          = {https://arxiv.org/abs/2007.00394}
}

@misc{sener2022assembly101,
  author       = {Sener, Fadime and Chatterjee, Dibyadip and Shelepov, Daniel and He, Kun and Singhania, Dipika and Wang, Robert and Yao, Angela},
  title        = {{Assembly101}: A Large-Scale Multi-View Video Dataset for Understanding Procedural Activities},
  year         = {2022},
  eprint       = {2203.14712},
  archivePrefix = {arXiv},
  url          = {https://arxiv.org/abs/2203.14712}
}

@InProceedings{madaan2023selfrefine,
  author       = {Madaan, Aman and Tandon, Niket and Gupta, Prakhar and Hallinan, Skyler and Gao, Luyu and Wiegreffe, Sarah and Alon, Uri and Dziri, Nouha and Prabhumoye, Shrimai and Yang, Yiming and Gupta, Shashank and Majumder, Bodhisattwa Prasad and Hermann, Katherine and Welleck, Sean and Yazdanbakhsh, Amir and Clark, Peter},
  title        = {{Self-Refine}: Iterative Refinement with Self-Feedback},
  year         = {2023},
  booktitle    = {Advances in Neural Information Processing Systems},
  volume       = {36},
  eprint       = {2303.17651},
  archivePrefix = {arXiv},
  url          = {https://dblp.org/rec/conf/nips/MadaanTGHGW0DPY23}
}

@InProceedings{shinn2023reflexion,
  author       = {Shinn, Noah and Cassano, Federico and Gopinath, Ashwin and Narasimhan, Karthik and Yao, Shunyu},
  title        = {{Reflexion}: Language Agents with Verbal Reinforcement Learning},
  year         = {2023},
  booktitle    = {Advances in Neural Information Processing Systems},
  volume       = {36},
  eprint       = {2303.11366},
  archivePrefix = {arXiv},
  url          = {https://dblp.org/rec/conf/nips/ShinnCGNY23}
}

@article{wang2023voyager,
  author       = {Wang, Guanzhi and Xie, Yuqi and Jiang, Yunfan and Mandlekar, Ajay and Xiao, Chaowei and Zhu, Yuke and Fan, Linxi and Anandkumar, Anima},
  title        = {{Voyager}: An Open-Ended Embodied Agent with Large Language Models},
  journal      = {Transactions on Machine Learning Research},
  year         = {2024},
  eprint       = {2305.16291},
  archivePrefix = {arXiv},
  url          = {https://openreview.net/forum?id=ehfRiF0R3a}
}

@InProceedings{yao2022react,
  author       = {Yao, Shunyu and Zhao, Jeffrey and Yu, Dian and Du, Nan and Shafran, Izhak and Narasimhan, Karthik and Cao, Yuan},
  title        = {{ReAct}: Synergizing Reasoning and Acting in Language Models},
  booktitle    = {International Conference on Learning Representations},
  year         = {2023},
  eprint       = {2210.03629},
  archivePrefix = {arXiv},
  url          = {https://openreview.net/forum?id=WE_vluYUL-X}
}

@InProceedings{wang2024codeact,
  author       = {Wang, Xingyao and Chen, Yangyi and Yuan, Lifan and Zhang, Yizhe and Li, Yunzhu and Peng, Hao and Ji, Heng},
  title        = {Executable Code Actions Elicit Better {LLM} Agents},
  year         = {2024},
  booktitle    = {Proceedings of the 41st International Conference on Machine Learning},
  series       = {Proceedings of Machine Learning Research},
  volume       = {235},
  pages        = {50208--50232},
  publisher    = {PMLR},
  eprint       = {2402.01030},
  archivePrefix = {arXiv},
  url          = {https://proceedings.mlr.press/v235/wang24h.html}
}

@misc{zhou2026articraft,
  author       = {Zhou, Matt and Li, Ruining and Lyu, Xiaoyang and Song, Zhaomou and Huang, Zhening and Zheng, Chuanxia and Rupprecht, Christian and Vedaldi, Andrea and Wu, Shangzhe},
  title        = {{Articraft}: An Agentic System for Scalable Articulated 3D Asset Generation},
  year         = {2026},
  eprint       = {2605.15187},
  archivePrefix = {arXiv},
  url          = {https://arxiv.org/abs/2605.15187}
}

@misc{pfaff2026scenesmith,
  author       = {Pfaff, Nicholas and Cohn, Thomas and Zakharov, Sergey and Cory, Rick and Tedrake, Russ},
  title        = {{SceneSmith}: Agentic Generation of Simulation-Ready Indoor Scenes},
  year         = {2026},
  eprint       = {2602.09153},
  archivePrefix = {arXiv},
  url          = {https://arxiv.org/abs/2602.09153}
}

@misc{chen2024assemblycomplete,
  author       = {Chen, Alan and Liu, Changliu},
  title        = {{AssemblyComplete}: 3D Combinatorial Construction with Deep Reinforcement Learning},
  year         = {2024},
  eprint       = {2410.15469},
  archivePrefix = {arXiv},
  url          = {https://arxiv.org/abs/2410.15469}
}

@misc{hong2023metagpt,
  author       = {Hong, Sirui and Zhuge, Mingchen and Chen, Jiaqi and Zheng, Xiawu and Cheng, Yuheng and Zhang, Ceyao and Wang, Jinlin and Wang, Zili and Yau, Steven Ka Shing and Lin, Zijuan and Zhou, Liyang and Ran, Chenyu and Xiao, Lingfeng and Wu, Chenglin and Schmidhuber, J{\"u}rgen},
  title        = {{MetaGPT}: Meta Programming for a Multi-Agent Collaborative Framework},
  year         = {2023},
  eprint       = {2308.00352},
  archivePrefix = {arXiv},
  url          = {https://arxiv.org/abs/2308.00352}
}

@InProceedings{gu2025manualvla,
  author       = {Gu, Chenyang and Liu, Jiaming and Chen, Hao and Huang, Runzhong and Wuwu, Qingpo and Li, Xiaoqi and Liu, Zhuoyang and Li, Ying and Zhang, Renrui and Jia, Peng and Heng, Pheng-Ann and Zhang, Shanghang},
  title        = {From Manuals to Actions: A Unified {VLA} Model for Chain-of-Thought Manual Generation and Robotic Manipulation},
  year         = {2026},
  booktitle    = {Proceedings of the IEEE/CVF Conference on Computer Vision and Pattern Recognition (CVPR)},
  eprint       = {2512.02013},
  archivePrefix = {arXiv},
  url          = {https://openaccess.thecvf.com/content/CVPR2026/html/Gu_From_Manuals_to_Actions_A_Unified_VLA_Model_for_Chain-of-Thought_CVPR_2026_paper.html}
}

@misc{jing2026assemlm,
  author       = {Jing, Zhi and Qiao, Jinbin and Lu, Ouyang and Ao, Jicong and Qiu, Shuang and Xu, Huazhe and Jiang, Yu-Gang and Bai, Chenjia},
  title        = {{AssemLM}: A Spatial Reasoning Multimodal Large Language Model for Robotic Assembly},
  year         = {2026},
  eprint       = {2604.08983},
  archivePrefix = {arXiv},
  url          = {https://arxiv.org/abs/2604.08983}
}

@misc{ma2026workbenchmark,
  author       = {Ma, Wenbo and Swoboda, Daniel and Tschesche, Matteo and Hofmann, Till},
  title        = {{WorkBenchMark}: A {LEGO}-Based Assembly Benchmark with an Assembly-by-Disassembly Baseline for the Smart Manufacturing League},
  year         = {2026},
  eprint       = {2606.19358},
  archivePrefix = {arXiv},
  url          = {https://arxiv.org/abs/2606.19358}
}

@misc{ni2026brickanything,
  author       = {Ni, Zhengyang and Yan, Feng and Guo, Yu and Wang, Fei},
  title        = {{BrickAnything}: Geometry-Conditioned Buildable Brick Generation with Structure-Aware Tokenization},
  year         = {2026},
  eprint       = {2605.26182},
  archivePrefix = {arXiv},
  url          = {https://arxiv.org/abs/2605.26182}
}

@misc{liu2026brickcomposer,
  author       = {Liu, Jiateng and Li, Bingxuan and Wang, Zhenhailong and Wang, Rushi and Hong, Kaiwen and Qian, Cheng and Liu, Jiayu and Zhang, Denghui and Driggs-Campbell, Katherine and Li, Manling and Ji, Heng},
  title        = {{Brick-Composer}: {MLLMs} Construct Everything from Building Blocks},
  year         = {2026},
  note         = {Manuscript},
  url          = {https://github.com/Lumos-Jiateng/Brick-Composer}
}

@misc{wang2026agenttrace,
  author       = {Wang, Zhaohui Geoffrey},
  title        = {{AgentTrace}: Causal Graph Tracing for Root Cause Analysis in Deployed Multi-Agent Systems},
  year         = {2026},
  eprint       = {2603.14688},
  archivePrefix = {arXiv},
  url          = {https://arxiv.org/abs/2603.14688}
}

@misc{kassianik2026fapo,
  author       = {Kassianik, Paul and Saglam, Baturay and Zhao, Huaibo and Nelson, Blaine and Vijay, Supriti and Priyanshu, Aman and Karbasi, Amin},
  title        = {{FAPO}: Fully Automated Prompt Optimization of Multi-Step {LLM} Pipelines},
  year         = {2026},
  eprint       = {2606.19605},
  archivePrefix = {arXiv},
  url          = {https://arxiv.org/abs/2606.19605}
}

@misc{lin2026harnessupdating,
  author       = {Lin, Minhua and Wu, Juncheng and Wang, Zijun and Shi, Zhan and Sang, Yisi and He, Bing and Liu, Zewen and Wei, Tianxin and Wu, Zongyu and Zhang, Zhiwei and Wang, Dakuo and Zhang, Xiang and Dumoulin, Benoit and Xie, Cihang and Zhou, Yuyin and Wang, Suhang and Lu, Hanqing},
  title        = {Harness Updating Is Not Harness Benefit: Disentangling Evolution Capabilities in Self-Evolving {LLM} Agents},
  year         = {2026},
  eprint       = {2605.30621},
  archivePrefix = {arXiv},
  url          = {https://arxiv.org/abs/2605.30621}
}

@article{liu2026skillforge,
  title={{SkillForge}: Forging Domain-Specific, Self-Evolving Agent Skills in Cloud Technical Support},
  author={Liu, Xingyan and Luo, Xiyue and Li, Linyu and Huang, Ganghong and Liu, Jianfeng and Qiao, Honglin},
  journal={arXiv preprint arXiv:2604.08618},
  year={2026},
  url={https://arxiv.org/abs/2604.08618}
}

@misc{claude,
  author={{Anthropic}},
  title={Claude Code},
  year={2025},
  url={https://github.com/anthropics/claude-code},
  note={Official product page: \url{https://claude.com/product/claude-code}. Accessed: 2026-09-02}
}

@misc{codex,
  author={{OpenAI}},
  title={Codex},
  year={2025},
  url={https://github.com/openai/codex},
  note={Official product page: \url{https://chatgpt.com/codex}. Accessed: 2026-09-02}
}

@misc{deepseekharness,
  author={{DeepSeek-AI}},
  title={DeepSeek Harness: Everything Is a Plugin},
  year={2026},
  url={https://github.com/deepseek-ai/deepseek-harness},
  note={Official product page: \url{https://www.deepseek.com/harness/}. Accessed: 2026-09-02}
}

@InProceedings{heo2023furniturebench,
  author       = {Heo, Minho and Lee, Youngwoon and Lee, Doohyun and Lim, Joseph J.},
  title        = {{FurnitureBench}: Reproducible Real-World Benchmark for Long-Horizon Complex Manipulation},
  booktitle    = {Robotics: Science and Systems},
  year         = {2023},
  eprint       = {2305.12821},
  archivePrefix = {arXiv},
  url          = {https://arxiv.org/abs/2305.12821}
}

@InProceedings{radford2021clip,
  author       = {Radford, Alec and Kim, Jong Wook and Hallacy, Chris and Ramesh, Aditya and Goh, Gabriel and Agarwal, Sandhini and Sastry, Girish and Askell, Amanda and Mishkin, Pamela and Clark, Jack and Krueger, Gretchen and Sutskever, Ilya},
  title        = {Learning Transferable Visual Models From Natural Language Supervision},
  booktitle    = {Proceedings of the 38th International Conference on Machine Learning},
  series       = {Proceedings of Machine Learning Research},
  volume       = {139},
  pages        = {8748--8763},
  year         = {2021},
  publisher    = {PMLR},
  eprint       = {2103.00020},
  archivePrefix = {arXiv},
  url          = {https://proceedings.mlr.press/v139/radford21a.html}
}

@article{oquab2024dinov2,
  author       = {Oquab, Maxime and Darcet, Timoth{\'e}e and Moutakanni, Th{\'e}o and Vo, Huy and Szafraniec, Marc and Khalidov, Vasil and Fernandez, Pierre and Haziza, Daniel and Massa, Francisco and El-Nouby, Alaaeldin and Assran, Mahmoud and Ballas, Nicolas and Galuba, Wojciech and Howes, Russell and Huang, Po-Yao and Li, Shang-Wen and Misra, Ishan and Rabbat, Michael and Sharma, Vasu and Synnaeve, Gabriel and Xu, Hu and J{\'e}gou, Herv{\'e} and Mairal, Julien and Labatut, Patrick and Joulin, Armand and Bojanowski, Piotr},
  title        = {{DINOv2}: Learning Robust Visual Features without Supervision},
  journal      = {Transactions on Machine Learning Research},
  year         = {2024},
  eprint       = {2304.07193},
  archivePrefix = {arXiv},
  url          = {https://openreview.net/forum?id=a68SUt6zFt}
}

@InProceedings{lin2024vqascore,
  author       = {Lin, Zhiqiu and Pathak, Deepak and Li, Baiqi and Li, Jiayao and Xia, Xide and Neubig, Graham and Zhang, Pengchuan and Ramanan, Deva},
  title        = {Evaluating Text-to-Visual Generation with Image-to-Text Generation},
  booktitle    = {European Conference on Computer Vision},
  pages        = {366--384},
  year         = {2024},
  eprint       = {2404.01291},
  archivePrefix = {arXiv},
  url          = {https://arxiv.org/abs/2404.01291}
}

@inproceedings{bolya2025perceptionencoder,
  author = {Bolya, Daniel and Huang, Po-Yao and Sun, Peize and Cho, Jang Hyun and Madotto, Andrea and Wei, Chen and Ma, Tengyu and Zhi, Jiale and Rajasegaran, Jathushan and Bangalath, Hanoona and Wang, Junke and Monteiro, Marco and Xu, Hu and Dong, Shiyu and Ravi, Nikhila and Li, Shang-Wen and Dollar, Piotr and Feichtenhofer, Christoph},
  title = {{Perception Encoder}: The Best Visual Embeddings Are Not at the Output of the Network},
  booktitle = {Advances in Neural Information Processing Systems},
  volume = {38},
  year = {2025},
  doi = {10.52202/085713-2036},
  url = {https://proceedings.neurips.cc/paper_files/paper/2025/hash/57bc0a850255e2041341bf74c7e2b9fa-Abstract-Conference.html}
}

@misc{openai2026gpt6astra,
  author       = {{OpenAI}},
  title        = {{GPT-6 Astra}},
  year         = {2026},
  howpublished = {OpenAI API model documentation},
  url          = {https://developers.openai.com/api/docs/models/gpt-6-astra},
  note         = {Accessed: 2026-09-26}
}

@misc{tschannen2025siglip2,
  author       = {Tschannen, Michael and Gritsenko, Alexey and Wang, Xiao and Naeem, Muhammad Ferjad and Alabdulmohsin, Ibrahim and Parthasarathy, Nikhil and Evans, Talfan and Beyer, Lucas and Xia, Ye and Mustafa, Basil and H{\'e}naff, Olivier and Harmsen, Jeremiah and Steiner, Andreas and Zhai, Xiaohua},
  title        = {{SigLIP 2}: Multilingual Vision-Language Encoders with Improved Semantic Understanding, Localization, and Dense Features},
  year         = {2025},
  eprint       = {2502.14786},
  archivePrefix = {arXiv},
  url          = {https://arxiv.org/abs/2502.14786}
}

@misc{qwen2026qwen36_27b,
  author = {{Qwen Team}},
  title = {{Qwen3.6-27B}},
  year = {2026},
  url = {https://huggingface.co/Qwen/Qwen3.6-27B},
  note = {Official model card. Accessed: 2026-09-26}
}
\bibliographystyle{iclr2027_conference}

\appendix
\clearpage
\section{Support Material of DailyBrick}
\label{app:compact-supplement}

\begin{figure}[h]
\centering
\includegraphics[width=0.98\linewidth]{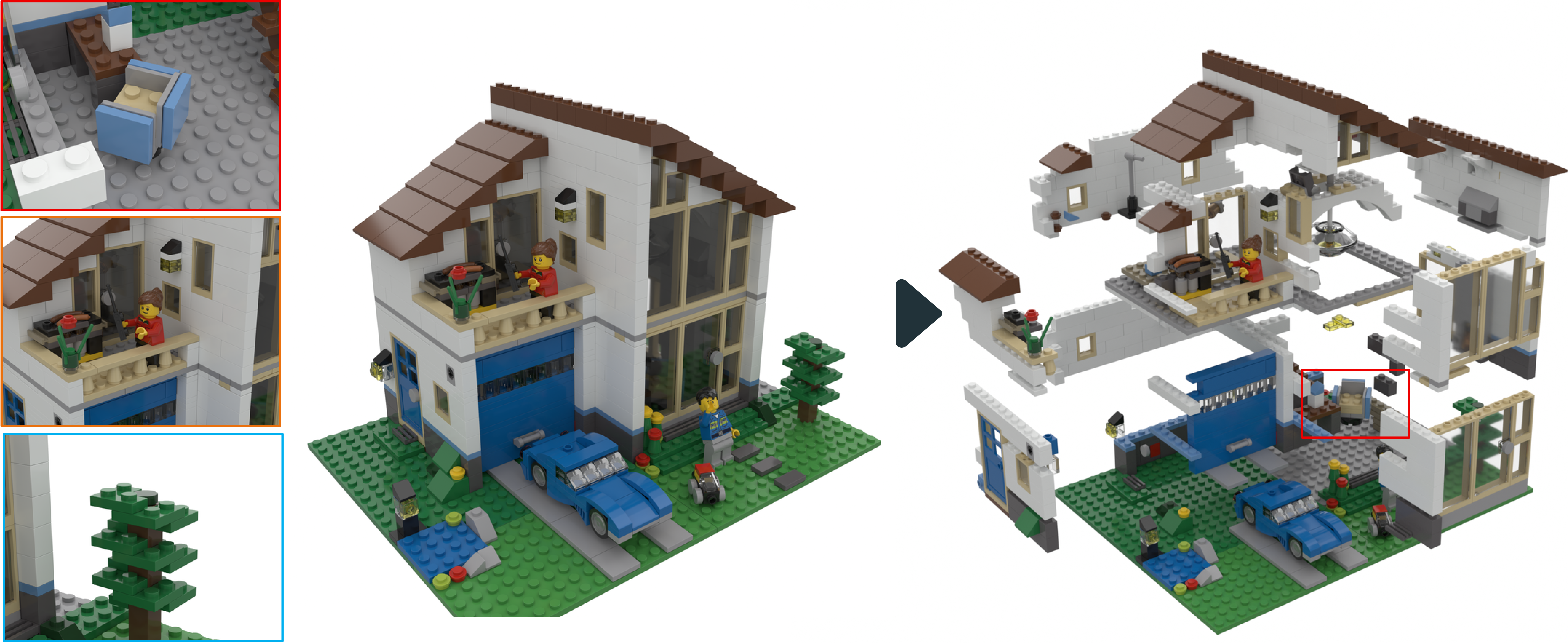}
\caption{\textbf{Staged planning and local assembly in DailyBrick.} The large house example illustrates how DailyBrick plans the main structure and its supports, then completes local components, including an interior sofa, an exterior balcony, and a decorative tree. Unlike approaches that fill a target shape with bricks while overlooking its interior, the staged construction represents both internal layout and external details.}
\label{fig:appendix-staged-assembly}
\end{figure}

\subsection{Additional Qualitative Results}
\label{app:qualitative-results}

\paragraph{Staged construction of a large model.}
Figure~\ref{fig:appendix-staged-assembly} illustrates how a large design is organized into a main structure and local components. The interior sofa, exterior balcony, and decorative tree represent different parts of the requested scene, while the house provides their spatial context. The complete and exploded views make both the internal arrangement and external details visible, illustrating the component-level organization used in staged construction.

\begin{figure}[h]
\centering
\includegraphics[width=0.5\linewidth]{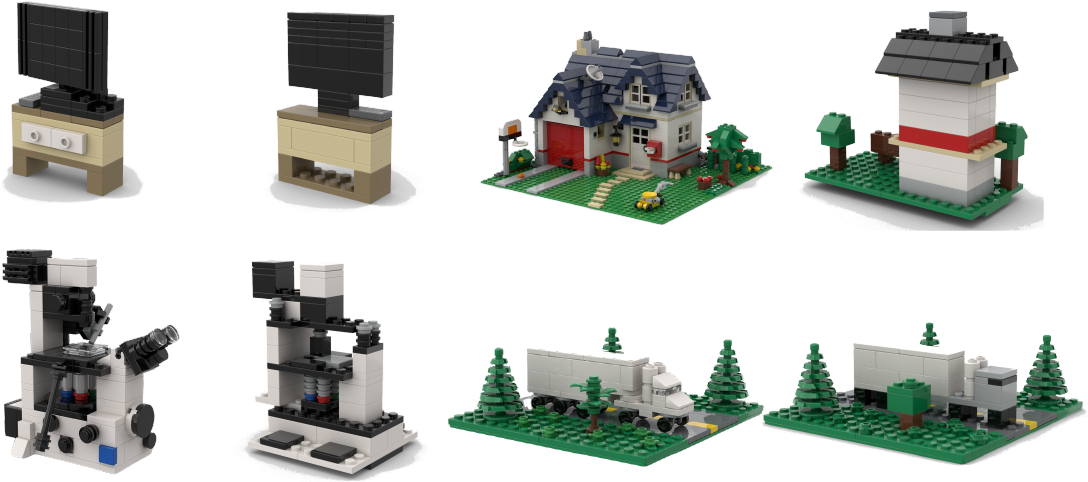}
\caption{\textbf{Generation with a restricted brick inventory.} Examples generated by DailyBrick (w/ Qwen3.6-27B) when specialized parts are excluded from the available brick inventory.}
\label{fig:appendix-brick-examples}
\end{figure}

\paragraph{Generation with a restricted inventory.}
Figure~\ref{fig:appendix-brick-examples} examines generation when specialized parts are excluded from the available inventory. This setting tests how DailyBrick expresses a design using the remaining parts. The examples provide a qualitative view of the resulting shapes and local construction details under the inventory restriction.

\begin{figure}[t]
\centering
\includegraphics[width=\linewidth]{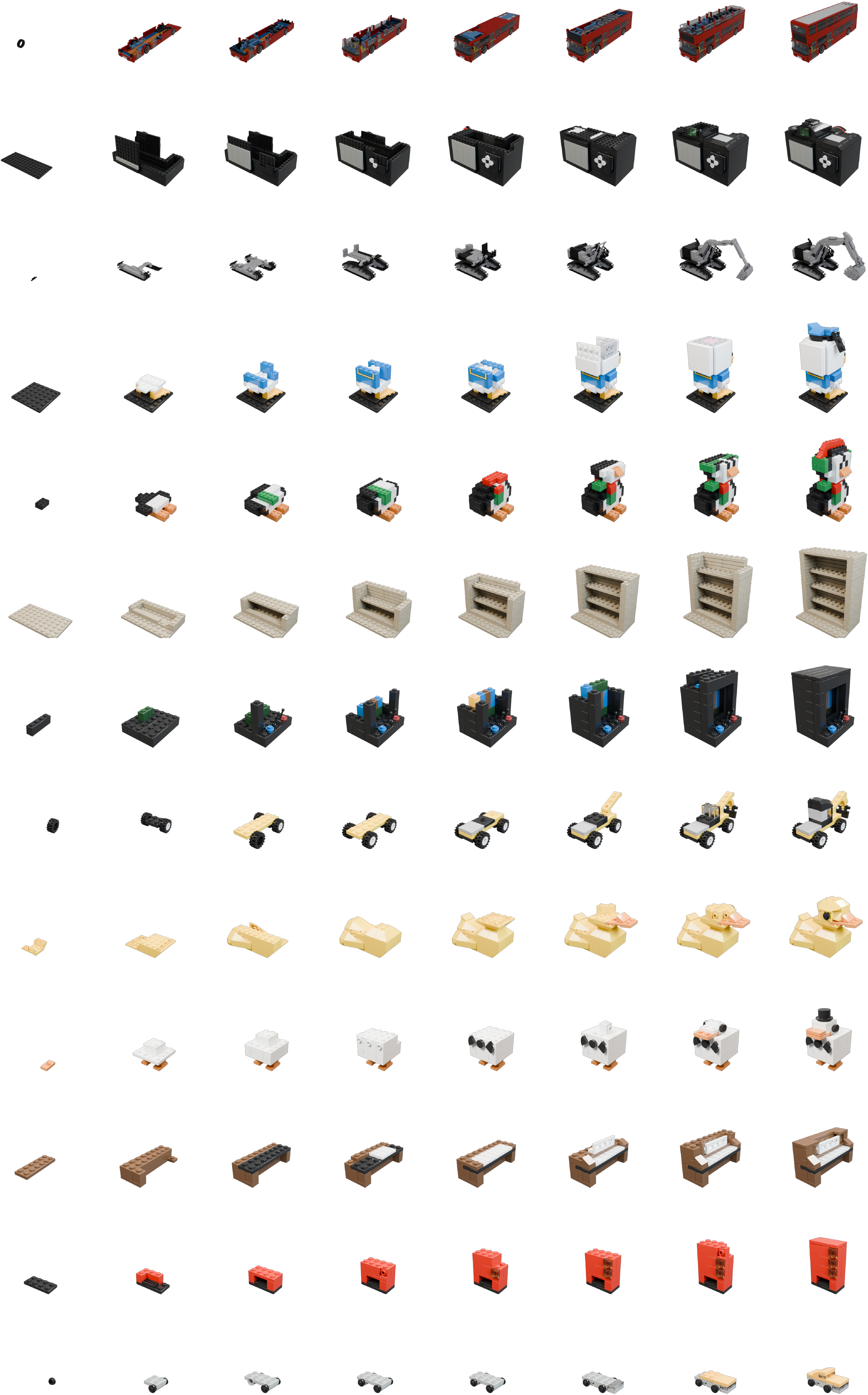}
\caption{\textbf{Step-by-Step Assembly Instructions from DailyBrick.} Intermediate assembly states generated by DailyBrick (w/ Qwen3.6-27B), shown in construction order from the initial placements to the completed model.}
\label{fig:appendix-dailybrick-assembly}
\end{figure}

\begin{figure}[t]
\centering
\includegraphics[width=\linewidth]{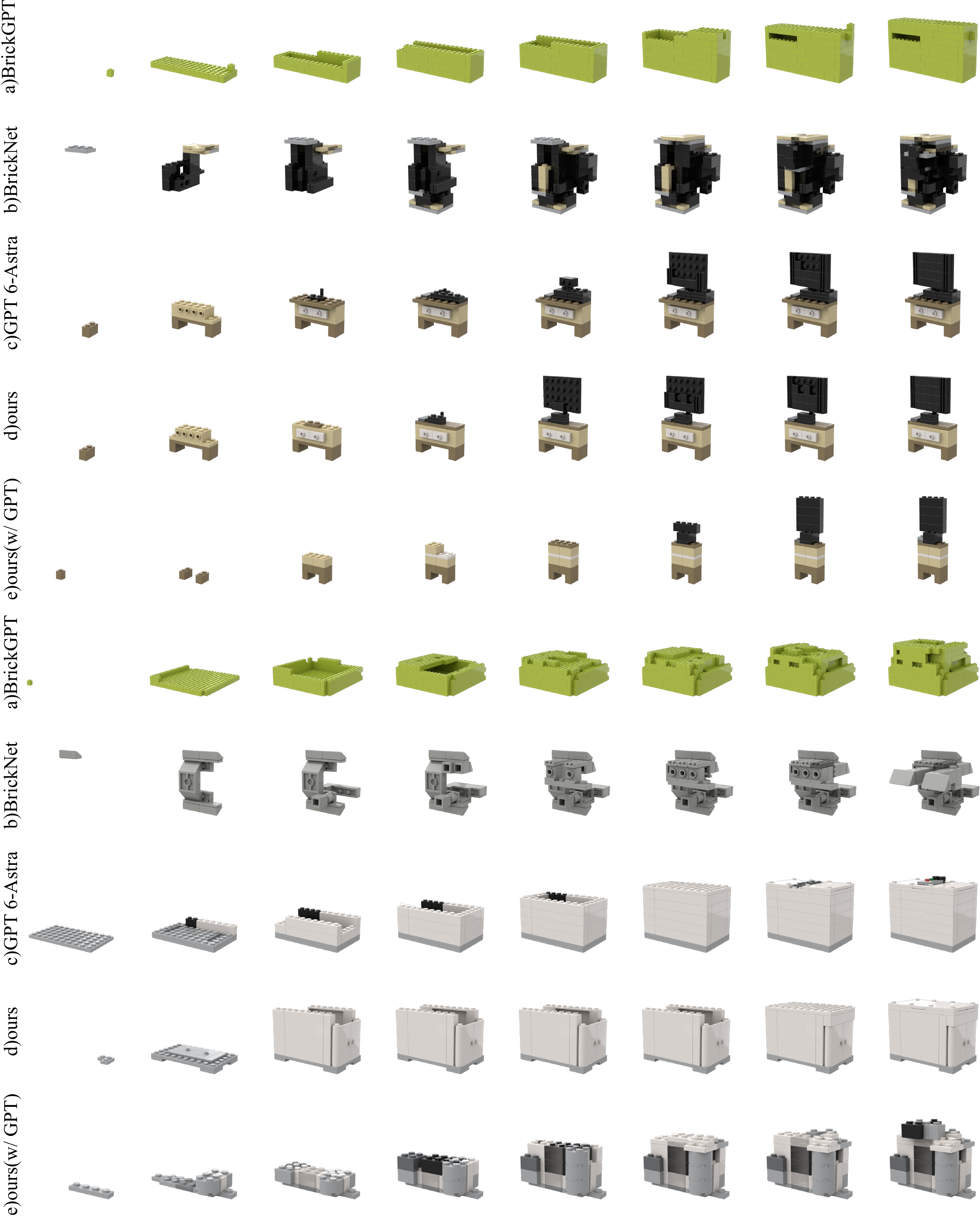}
\caption{\textbf{Step-by-Step Assembly Instruction Visualization of LEGO Models}. The television cabinet and white cabinet are each shown using five methods: BrickGPT, BrickNet, GPT 6-Astra, ours (with Qwen3.6-27B) and ours w/ GPT 6-Astra. Each row presents eight stages of one model: the first assembly step, six uniformly sampled intermediate steps, and the completed model.}
\label{fig:appendix-selected-images}
\end{figure}

\paragraph{Step-by-step assembly instructions.}
Figure~\ref{fig:appendix-dailybrick-assembly} presents step-by-step assembly instructions generated by DailyBrick with Qwen3.6-27B, showing how successive brick placements build up a model from partial assemblies to its completed form. Figure~\ref{fig:appendix-selected-images} further compares intermediate assemblies from five methods on two furniture examples. Together, these visualizations show how DailyBrick's output specifies a construction process in addition to a final model, and allow comparison with the sequences produced by the other methods.

\newpage

\newpage

\subsection{Additional Method Details}
\label{app:method-details}

\subsubsection{Stage Coordination and Intermediate Outputs}

DailyBrick divides generation into request interpretation, abstraction and shape planning, database grounding, semantic assembly planning, and SDK construction. Each LLM-governed stage follows a Planner--Designer--Critic loop with a prescribed output schema. The Planner uses the stage requirements and upstream outputs to direct the local task. The Designer produces a candidate, and the Critic reviews it together with deterministic validation results. Each round records the planner action, designer output, validation results, critic feedback, and acceptance decision. These records provide evidence for subsequent review and refinement.

\subsubsection{Construction Programs and Partial Assembly Skills}

An assembly program expresses the design through construction SDK commands. Available operations include frames, panels, and shells, as well as reusable assembly patterns such as stepped roofs, wheel pairs, hinge pairs, and support cores. Partial assembly skills allow local components to be specified through these operations and expanded into explicit brick instances and build steps. Before execution, the program validator checks the command format, permitted operations, and coverage of the expected semantic parts. When required, it also checks whether the program incorporates the retrieved construction techniques and runtime constraints. The resulting construction sequence provides the common basis for the final model, instruction views, and construction checks.

\subsubsection{Verify Replay and Construction Evidence}

Verify replay checks the recorded construction sequence against the partial assembly at each step. The construction checks examine instance identities, geometric intersections, support relationships, orientations, attachments, and declared dependencies. Step reports record the relevant instance identifiers, support sources, and individual check outcomes. When a check fails, the reported action and violated constraint provide evidence for error localization. These checks assess construction feasibility within the declared geometric and connection constraints. Robot grasping, continuous motion, and force-dependent behavior require additional downstream evaluation.

\subsubsection{Self Meta-Reviewer and Targeted Refinement}

The Self Meta-Reviewer uses construction reports and stage-level evidence to locate errors and select candidate revision targets. The evidence includes the affected field, defect type, severity, and associated stage outputs. Missing semantic components can motivate a revision of the component plan, color failures can point to database grounding, and program or layout failures can require changes to SDK construction. Within a run, revisions modify the generated program or stage outputs while keeping the configuration \(\Gamma\) fixed. The affected downstream outputs are regenerated and checked again before acceptance. If the evidence is insufficient to select a revision target, automatic refinement is deferred. This subsection describes repairs within a run; updates to prompts, skills, or SDK definitions belong to the later-configuration update process described in the main paper.

\end{document}